\documentclass[letterpaper, 10 pt, conference]{ieeeconf}  

\IEEEoverridecommandlockouts                              

\usepackage{graphicx}
\usepackage{url}  
\usepackage{xcolor}

\usepackage{amsmath}

\usepackage{caption}
\usepackage{subcaption}

\usepackage{fancyhdr,lipsum} 

\fancypagestyle{firstpage}{
  \fancyhf{}

}

\title{\LARGE \bf
An Analysis of State-of-the-art Activation Functions \\ For Supervised Deep Neural Network 
}

\author{Anh~Nguyen$^{1}$, 
             Khoa~Pham$^{2}$, 
             Dat~Ngo$^{1}$, 
             Thanh~Ngo$^{3}$,
             Lam~Pham$^{4}$ 
\thanks{$^1$A. Nguyen, D. Ngo, are with Department of Electronics, Ho Chi Minh City University of Technology, Vietnam National University - Ho Chi Minh, Viet Nam.}%
\thanks{$^2$K. Pham is with Department of Computer and Communication Engineering, University of Technology and Education HCMC, Viet Nam.} %
\thanks{$^3$T. Ngo is with Department of Electrical Engineering, Da Nang University of Science and  Technology, Vietnam.}
\thanks{$^4$L. Pham is with Center for Digital Safety \& Security, Austrian Institute of Technology, Austria.}
}

\begin{document}

\maketitle
\thispagestyle{firstpage}
\pagestyle{empty}

\begin{abstract}

This paper provides an analysis of state-of-the-art activation functions with respect to supervised classification of deep neural network. These activation functions comprise of Rectified Linear Units (ReLU), Exponential Linear Unit (ELU), Scaled Exponential Linear Unit (SELU), Gaussian Error Linear Unit (GELU), and the Inverse Square Root Linear Unit (ISRLU). 
To evaluate, experiments over two deep learning network architectures integrating these activation functions are conducted. 
The first model, basing on Multilayer Perceptron (MLP), is evaluated with MNIST dataset to perform these activation functions.
Meanwhile, the second model, likely VGGish-based architecture, is applied for Acoustic Scene Classification (ASC) Task 1A in DCASE 2018 challenge, thus evaluate whether these activation functions work well in different datasets as well as different network architectures. 


\end{abstract}
\section{Introduction}
\subsection{The Role of Activation Function in Deep Neural Network}
\begin{figure}[th]
    \centering
    \includegraphics[width=.6\linewidth,height=5.7cm]{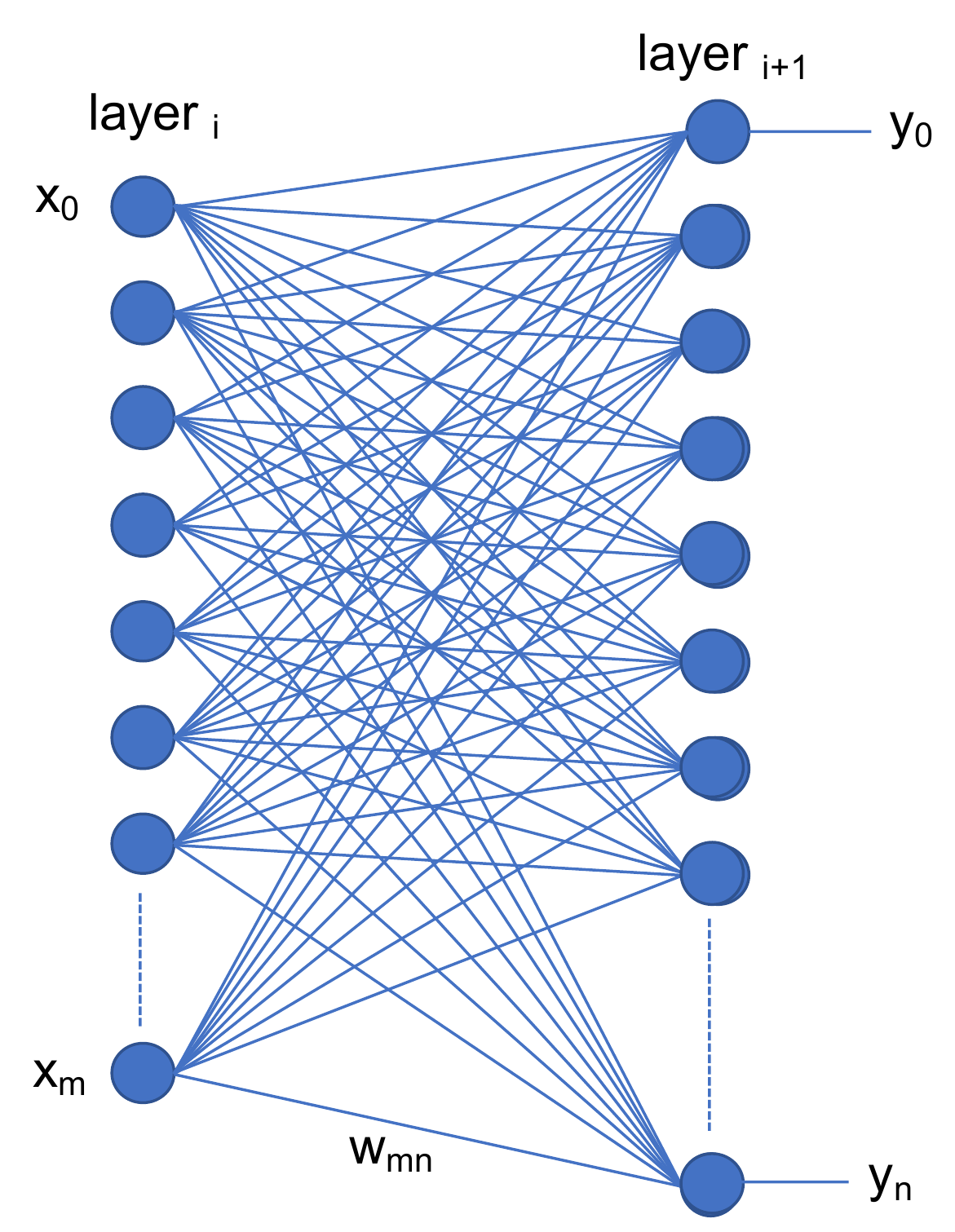}
	\caption{A fully-connected layer}
    \label{fig:F1}
\end{figure}
Consider a simple fully-connected layer as shown in Fig. \ref{fig:F1}, the output column vector $\mathbf{y^{T}} = [y_1, y_2, ..., y_n]$ is computed by
\begin{equation}
       \label{eq: connection}
       \mathbf{y} = f \left( \mathbf{W} \mathbf{x} + \mathbf{b} \right)
\end{equation}

where $\mathbf{x^{T}} = [x_1, x_2, ..., x_m]$ is the input column vector, matrix $\mathbf{W}$ and bias $\mathbf{b^{T}} = [b_1, b_2, ..., b_n]$ are trainable parameters, and \(f\) is activation function selected. 
From \eqref{eq: connection}, it can be seen that a simple fully-connected layer is separated into two mathematical parts in order.
The first part describes a linear mapping, shown by a product between input vector \(\mathbf{x}\) and trainable parameters (matrix  $\mathbf{W}$), thus added by the bias vector \(\mathbf{b}\).
The second part, called activation function \(f\), takes the role of non-linear mapping to output \(\mathbf{y_{n}}\).
The combination of linear and non-linear mapping mentioned helps the mathematical model operate as a real neural network \cite{mohammadzaheri2009combination}

\subsection{Backward \& Forward Propagation in Deep Neural Network}
During training process, input is fed into the network firstly.
The input thus goes through network layers, experiences various linear and non-linear mappings, eventually results final output, referred to as inference output.
This process is called \textit{Forward Propagation} that shows a direction from input to output.
Next, the difference between inference output and expectation (or ground truth) is performed by a loss function (or cost function).
Minimizing the loss function is conducted by applying an optimization algorithm, namely \textit{Gradient Descent}.
By using \textit{Gradient Descent} optimization, any trainable parameter $w_t$ at certain time $t$ in $\mathbf{W}$ mentioned in \eqref{eq: connection} is updated by 
\begin{equation}
       \label{eq: update}
      w_t = w_{t-1} - \eta \nabla f(w_{t-1})
\end{equation}
where \(\eta\) is learning rate (noting that the learning rate is normally constant in certain epoch), \(f(w_{t-1})\) is multi-variable function as mentioned in \eqref{eq: connection}, \(\nabla\) is derivative, and \(\nabla f(w_{t-1})\) is the gradient value.
The subtraction means that a trainable parameter is updated in the opposite direction of gradient vector.
In particular, if gradient value is negative, the trainable parameter is added with the absolute of \(\nabla f(w)\). 
if gradient is positive, the trainable parameter is subtracted by the absolute of \(\nabla f(w)\). 
The nearer trainable parameters are to the output layer, the sooner they are updated. 
Therefore, the process of updating trainable parameters, referred to as \textit{Backward Propagation}, shows the direction from output to input.
\subsection{State-of-the-art Activation Functions Evaluated}
\label{act_func}

\begin{figure}
    \centering
    \begin{subfigure}[b]{.45\linewidth}
        \centering
        \includegraphics[width=\linewidth]{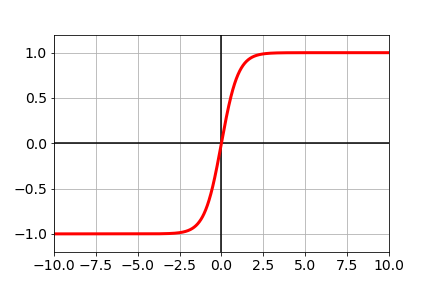}
	    \caption{Tanh function}
        \label{fig: Tanh}
    \end{subfigure}
    \hfill
    \begin{subfigure}[b]{.45\linewidth}
        \centering
        \includegraphics[width=\linewidth]{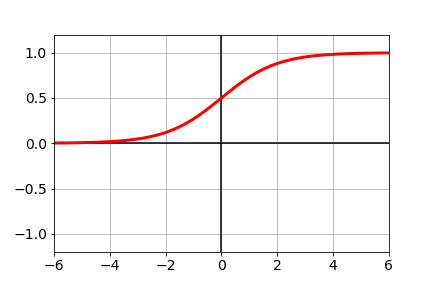}
        \caption{Sigmoid function}
        \label{fig: Sigmoid}
    \end{subfigure}
    \caption{Tanh and Sigmoid functions}
    \label{fig: activation function}
\end{figure}
This section analyses state-of-the-art activation functions used by a wide range of network architectures, comprising of ReLU, ELU, SELU, GELU and ISRLU.
Before going through these activation functions, we firstly analyse conventional activation functions: Tanh \cite{kalman1992tanh} and Sigmoid \cite{han1995influence} activation functions. 
As Tanh function shown in Fig.~\ref{fig: Tanh}, it helps to map the product of inputs and trainable parameters to a range from -1 to 1.
This function shows saturation regions where input makes output near -1 or 1.
When input makes output drop in these regions, \textit{Back Propagation} process shows a potential problem that the gradient is too small, named \textit{vanishing gradients}.
When a network suffers from \textit{vanishing gradients}, trainable parameters will not be adjusted, result in stopping learning.
As regards Sigmoid activation function, it has the same shape of Tanh function, and shows the same problem of \textit{vanishing gradients} when output drops in saturation regions near 0 or 1.

\textbf{ReLU} \cite{krizhevsky2017imagenet}: As formula of ReLU activation function and its derivative shown by,
\begin{equation}
       \label{eq: relu}
       f(x)=\begin{cases}
     x, & \text{if $x>0$}.\\
     0, & \text{if $x\leq0$}.
   \end{cases}
\end{equation}
this function shows simple formula and its gradient is also easy to get, as shown by
\begin{equation}
       \label{eq: relu_dev}
       f'(x)=\begin{cases}
     1, & \text{if $x>0$}.\\
     0, & \text{if $x\leq0$}.
   \end{cases}
\end{equation}
As output  \(f(x)\)  is zero for values smaller than zero, otherwise one, it explains why ReLU prevents \textit{vanished gradients}. 
However, a potential issue can occur with ReLU that is called \textit{dead state} when it is stuck at the left side of zero.
If the issue happens to many nodes inside a network, it makes the network low performance.
Furthermore, since the sign of gradient for all neurons is the same with ReLU, all the weights of the layer either increase or decrease. 
However, the ideal gradient update is that some trainable parameters may increase and others may decrease. 
It can be seen that ReLU helps to reduce cost of computation, speed up training process, prevent \textit{vanishing gradients}, but it shows the other potential problems recently mentioned.

\textbf{ELU} \cite{clevert2015fast}: It can be seen that Tanh or Sigmoid activation function have a two-sided saturation.
That means the activation function may saturate in both the positive and the negative direction. 
In contrast, ReLU activation function shows one-sided saturations.
Up to now, it seems that there are two risks: \textit{dead state} by dying neurons with ReLU or a risk for \textit{vanishing gradients} with Sigmoid or Tanh.
To tackle this issue, ELU activation function is developed by
\begin{equation}
       \label{eq: elu}
       f(x)=\begin{cases}
     x, & \text{if $x>0$}.\\
     \alpha(e^{x} - 1), & \text{if $x\leq0$}.
   \end{cases}
\end{equation}

\begin{equation}
       \label{eq: elu_dev}
       f'(x)=\begin{cases}
     1, & \text{if $x>0$}.\\
     \alpha e^{x}, & \text{if $x\leq0$}.
   \end{cases}
\end{equation}
where \(\alpha\) is constant that normally set to 1. 
It can be seen that ELU is very exactly similar to ReLu with respect to positive inputs. 
However, when input is negative, ELU function results in a negative output ruled by an exponent function.
Since, ELU produces a negative value instead of 0 as ReLu function, the problem is that weights updated only in one direction cannot occur.
By using exponent function when input is negative, there is now a slope on the left side of ELU. 
Although the slope is usually tiny (e.g. 0.01), but it is there, makes learning happening always and the neural node cannot die.
As a result, it helps to solve both \textit{vanishing gradients} and \textit{dead state} problems.

\textbf{SELU} \cite{selu}: In terms of neural networks and training process, normalization can occur in three different stages. 
Firstly, input is normalized before feeding into network model. 
Secondly, it is batch normalization where the batch of input are normalized at the certain layer of the network model.
Finally, it is called internal normalization or self normalization, and SELU is developed for this purpose. 
If input and output of all layers in a network model follow normal distribution (\(\mu = 0\),  \(\sigma = 1\)), it is helpful to speed up training process by quickly converging.
Inspire from this idea, SELU activation function, shown as \eqref{eq: selu}, is designed to be able to make output into a normal distribution. 
In the other words, the function operation helps to map the mean and variance of output at any layer in the network close to normal distribution.

\begin{equation}
       \label{eq: selu}
       f(x)= \begin{cases}
    \lambda x, & \text{if $x>0$}.\\
     \lambda \alpha (e^{x} - 1), & \text{if $x\leq0$}.
   \end{cases}
\end{equation}

\begin{equation}
       \label{eq: selu_dev}
       f'(x)= \begin{cases}
     \lambda, & \text{if $x>0$}.\\
     \lambda \alpha e^{x}, & \text{if $x\leq0$}.
   \end{cases}
\end{equation}
\(\lambda\) is  called scale constant.  \(\lambda\) and \(\alpha\) is set to 1.0507 and 1.6732 that helps to make output distribution nearer normal distribution~\cite{selu}.
However, in order to use SELU, it requires some conditions (a) the neural network consists only a stack of dense layers, (b) all the hidden layers use SELU activation function, (c) the input features should be standardized, (d) the hidden layers weight must be initialized with LeCun normal initialization, and (e) the network must be sequential.
The Lecun normal initialization is describe by
\begin{equation}
       \label{eq: LeCun}
       w_{n} \sim \mathcal{N}(\mu=0, \sigma= \sqrt{1/n})
\end{equation}
where \(n\) is the number of trainable parameters of previous layer.
In summary, SELU is similar to ReLU that enables deep neural networks to tackle \textit{vanishing gradients}.
In contrast to RELU, SELU cannot die as they have negative side (same as ELU).
In case that conditions of using SELU are satisfied, it is beneficial to network to learn faster and be better than other activation functions mentioned.

\textbf{GELU} \cite{hendrycks2016gaussian}: As analyse mentioned above, activations such as ReLU, ELU and SELU enable network become faster and better convergence than that in Sigmoid or Tanh.
However, these cannot cover Dropout regularization \cite{srivastava2014dropout} that randomly multiplies a few activation functions by 0 at certain nodes in layers.
Inspiration from an idea that an activation function would have both these advantages, GELU activation function~\cite{gelu} is developed. 
Mathematically, if ELU is used, the output of this function is thus multiplied by a zero-one mask. 
This means that a dropout technique is applied after ELU function.
The zero-one mask is stochastically determined, also dependent upon the input \(x\), and simulated by Bernoulli distribution \(\Theta(x)\) where \(\Theta(x) = P(X \leq x), X \sim \mathcal{N}(0, 1)\).
With this idea, the GELU is approximated by
 \begin{equation}
       \label{eq: gelu}
       f(x) = 0.5 x (1 + tanh[\sqrt{2/\pi}(x + 0.044715 x^{3})]) 
\end{equation}

Thus, derivative of GELU function used for back propagation is presented by
 \begin{equation}
    \label{eq: gelu_dev}
    \begin{split}
        f'(x) &= 0.5tanh(0.0356774x^{3}+0.797885x)\\
        &+ 0.5 + (0.0535161x^{3}+0.398942x)\\ 
        &\times cosh^{-2}(0.0356774x^3+0.797885x)
    \end{split}
\end{equation}

\textbf{ISRLU} \cite{isrlu}: As regards ELU activation function, it is beneficial to solve \textit{vanishing gradients}. 
Furthermore, as this function results in both negative and positive output for both sides of zero, it is useful for the network to shift mean to close zero, leading to a near distribution that speeds up training process and converging.
However, the exponent function on the left side of zero highly cost computation.
To further speed up learning, Inverse Square Root Linear Unit (ISRLU)~\cite{isrlu} is proposed by,

\begin{equation}
       \label{eq: isrlu}
       f(x)= \begin{cases}
     x, & \text{if $x\geq0$}.\\
     x(\frac{1}{\sqrt{1 + \alpha x^{2}}}), & \text{if $x<0$}.
   \end{cases}
\end{equation}

\begin{equation}
       \label{eq: isrlu_dev}
       f'(x)= \begin{cases}
     1, & \text{if $x\geq0$}.\\
     (\frac{1}{\sqrt{1 + \alpha x^{2}}})^{3}, & \text{if $x<0$}.
   \end{cases}
\end{equation}

where constant \(\alpha\) is normally ranged from 1 to 3.
As can be seen that the exponent function is replaced by a polynomial that is useful to reduce cost of computation.

\section{Evaluation Activation Functions with MLP and MNIST dataset}
To evaluate these activation functions mentioned, we firstly propose a neural network, basing on Multilayer Perceptron (MLP) \cite{Goodfellow-et-al-2016}, and verify the network with MNIST dataset \cite{lecun1998gradient}.
Although a lot of suggestions were provided to configure a MLP based network, there is no magic formula to select the optimum number of hidden layers or hidden neurons in each layer \cite{ba2014deep} in terms of the MLP architecture.
Therefore, we present strategies to configure a MLP based network below.
 
\subsection{Decision of the number of hidden layers}
Given by MNIST dataset \cite{lecun1998gradient}, while the input and output size are fixed to 784 (an MNIST image size is $28\times28$ is flatten into 784-dimensional vector) and 10 (the number of categories form 0 to 9 classified) respectively, the minimum number of hidden layers should be larger.
This is explained by:

\begin{itemize}
\item Without hidden layers, the network only is capable of representing linearly separable functions or decisions. 
\item The network with only one hidden layer approximates any function that contains a continuous mapping from one finite space to another.
\item Two-hidden-layer networks can represent an arbitrary decision boundary to arbitrary accuracy with rational activation functions and can approximate any smooth mapping to any accuracy.
\item More-than-two-hidden-layer networks can learn complex representations (sort of automatic feature engineering).  
\end{itemize}

Therefore, three hidden layers are decided to do experiments in this paper. 
Although three hidden layers may or may not enough to learn MNIST dataset, it is useful to speed up training process due to small network size.
Selecting low volume of trainable parameters also helps network more general, thus evaluates the effect of mentioned activation functions better.

\subsection{The number of nodes in each hidden layer}
That is hard questions to indicate suitable number of nodes in each layer. This paper proposes some strategies below,
\begin{itemize}
\item The number of neural nodes in each hidden layer should be a power of two.
\item The selected number of nodes are evaluated by conducting experiments.
\end{itemize}

\subsection{Experimental setting}
MNIST dataset is randomly separated into three subsets. The first subset is used for training. During training, the second subset, referred to as Evaluation subset (Eva.), is used to evaluate model, thus store the best model.
Finally, the third subset, referred to as Test subset (Test), is used to evaluate the best model obtained from the training process.

In terms of network parameters, the initial trainable parameters are selected by Uniform distribution with scale set to 0.1 and learning rate is set to 0.001. These values are be fixed during evaluating the number of nodes in each hidden layer. Some other settings are decided below,
\begin{itemize}
\item Using ReLU as the activation function at hidden layers. 
\item Using Softmax function at the final layer to perform output probability. 
\item Apply Adam algorithm \cite{kingma2014adam} for optimization and Cross-entropy is used for loss function as Eq. (\ref{eq:loss_func})
\begin{equation}
    \label{eq:loss_func}
    \text{Loss}_{\text{EN}}(\Theta) = -\frac{1}{\text{N}}\sum_{n=1}^{N}\mathbf{y_n} log \left\{\mathbf{\hat{y}_{n}}(\Theta) \right\} + \frac{\lambda}{2} ||\Theta||_{2}^{2}
\end{equation}
where \(\Theta\) is all parameters, constant \(\lambda\) is set initially to $0.0001$,  both batch size $N$ and the epoch number are set to 100, and  $\mathbf{y_{n}}$ and $\mathbf{\hat{y}_{n}}$  denote expected and predicted results.
\end{itemize}

\subsection{Experiment Results and Discussion}
\label{mpl}

As shown in Table \ref{table:MLP}, we evaluate 9 different MLP based network architectures and the network architecture of 784-2048-2048-512-10 shows the best performance of 97.5\% and 96.4\% over Evaluation (Eva.) and Test (Test) subsets, respectively.
Thus, this network configuration is selected to evaluate activation functions.
\begin{table}[th!]
    \caption{Different MLP configurations with the highest accuracy in \textbf{bold}} 
    \centering
    \scalebox{0.9}{
    \begin{tabular}{c c c c c c} 
        \hline 
	    \textbf{Input } & \textbf{Hidden} & \textbf{Hidden} & \textbf{Hidden} & \textbf{Output}  & \textbf{Acc. \%} \\[0.5ex] 
	    
	    \textbf{layer} & \textbf{layer 01} & \textbf{layer 02} & \textbf{layer 03} & \textbf{layer}  & \textbf{(Eva./Test) } \\[0.5ex] 

        \hline 
         784               & 2048   &1024                &512  &10    & 96.6/96.0\\
        784               & 1024   &512                &256   &10 & 94.4/94.4\\
        784               & 512   &256                &128   &10 & 92.4/92.6\\
        784               & 256   &128                &60  &10 & 89.9/90.4\\
       \hline 
        784               & 512   &512                &512   &10  & 94.2/94.5\\
        784               & 1024   &1024                &256   &10  & 95.2/95.3\\
        784               & 1024   &1024                &512   &10   & 95.6/95.7\\
        784               & 1024   &1024                &1024   &10  & 9.65/9.61\\
        \textbf{784}               & \textbf{2048}   &\textbf{2048}                &\textbf{512}   &\textbf{10}  & \textbf{97.5/96.4}\\        
        \hline 
    \end{tabular} 
    }
    \label{table:MLP} 
\end{table}
\begin{table}[th]
    \caption{Activation functions' performance with the highest accuracy and lowest time consumption in \textbf{bold}} 
    \centering
    \scalebox{0.9}{
    \begin{tabular}{c c c c c c c} 
        \hline 
	    \textbf{Activation}  & \textbf{Eva./Test}  & \textbf{Training Time/Epoch} \\[0.5ex] 
	    \textbf{Function}  & \textbf{Acc. (\%)}  & \textbf{(second)} \\[0.5ex] 

        \hline 
        RELU                  & 97.5/96.4            &\textbf{11.86}   \\
        ELU                      &97.3/96.5             &12.14     \\
        SELU                    &\textbf{99.0/97.0}             &12.79      \\
        GELU                  &97.1/96.6              &82.37     \\
        ISRLU                   &97.9/96.8      &23.98    \\

        \hline 
    \end{tabular}
    }
    \label{table:act_func} 
\end{table}
By using the MLP network architecture of 784-2048-2048-512-10 selected and keeping setting parameters of learning rate, Epoch number and Batch size, ReLU function is replaced  by ELU, SELU, GELU, and ISRLU functions to evaluate in the order.
As results obtained shown in Table \ref{table:act_func}, it can be seen that four activation functions analysed perform better than ReLU. 
This can be explained that these four activation functions result certain value when input is negative, prevent both the \textit{vanishing gradients} and the \textit{dead state}.
Meanwhile ReLU may drop in the \textit{dead state} when input is negative. 
In terms of training time, ELU and SELU consume little more time per training epoch compared with ReLU.
GELU and ISRLU, by contrast, cost more time.

\section{Evaluate Activation Functions with VGGish network and DCASE 2018 Task 1A}
To evaluate whether these activation functions also work well on different datasets and network architectures with respect to a certain application, we further conduct experiments over a Convolutional Neural Network (CNN) with DCASE 2018 Task 1A dataset~\cite{Mesaros2018_DCASE}. 

\subsection{DCASE 2018 Task 1A dataset}
DCASE 2018 Task 1A~\cite{Mesaros2018_DCASE} defines a Acoustic Scene Classification task, one of two major researchs of \textit{Machine Hearing} \cite{lyon2017human}. In particular, the challenge is to classify 10-second audio recordings into 10 sound scene context as shown in Table \ref{table:DCASE2018_dataset}.
As evaluation set has not published, development set is separated into two subsets, referred to as training set and test set, for training and testing processes respectively.

\subsection{Feature Engineering Extraction}
Firstly, the draw audio recordings are transformed into spectrograms where both temporal and frequency information are well presented. Next, the entire spectrograms are splitted into smaller patches of $128\times128$ that are suitable for back-end classification. 
In this paper, we apply log-mel for spectrogram transformation and use Librosa toolbox \cite{mcfee2015librosa} to implement with setting parameters shown in Table \ref{table:spectrogram}.
\begin{table}[th!]
    \caption{DCASE2018 Task 1A dataset} 
    \centering
    \scalebox{0.9}{
    \begin{tabular}{c c c } 
        \hline 
	    \textbf{Categories}  & \textbf{Training set} & \textbf{Test set} \\ [0.5ex] 
        \hline 
        Airport                  & 599               & 265   \\
        Bus                      & 622               & 242     \\
        Metro                    & 603               & 261   \\
        Metro Station            & 605               & 259    \\
        Park                     & 622               & 242   \\
        Public Station           & 648               & 216   \\
        Shopping Mall            & 585               & 279   \\
        Pedestrian Street        & 617               & 247   \\
        Traffic Street           & 618               & 246   \\
        Tram                     & 603               & 261   \\
        \hline 
        \textbf{Total files}              & \textbf{6122}              & \textbf{2518}   \\
        \hline 

    \end{tabular}
    }
    \label{table:DCASE2018_dataset} 
\end{table}

\begin{table}[th]
    \caption{Spectrogram parameters} 
    \centering
    \scalebox{0.9}{
    \begin{tabular}{l c} 
        \hline 
            \textbf{Parameters}   &  \textbf{Setting}  \\
        \hline 
        FFT Number & 2048\\
        Hop Size & 256\\
        Window Size & 1024\\
        Log-mel Filter Number & 128\\
        Patch Size & $128\times128$\\
       \hline 
    \end{tabular}
    }
    \label{table:spectrogram} 
\end{table}

\subsection{Back-end CNN based Classifier}
As regards back-end classification, a CNN based network architecture, likely VGG-9 \cite{simonyan2014very}, is proposed as shown in Table \ref{table:CDNN}.
It can be seen that the CNN based model comprises of two main parts, CNN and DNN in the order. 
The CNN part is performed by covolutional layer (Cv [kernel size]), Rectify Liner Unit (ReLU), Batch normalization (BN), Average pooling (AP [kernel size], Drop out (Dr (dropped percentage)), and Global average pooling (GAP).
Meanwhile, DNN part comprises of three fully connected layers (FC) followed by Rectify Linear Unit and Drop out. At the final fully connected layer, Softmax is used to perform probability of 10 categories. 

\begin{table}[htb]
    \caption{CNN based network architecture} 
    \centering
    \scalebox{0.9}{

    \begin{tabular}{l c} 
        \hline 
            \textbf{Network architecture}   &  \textbf{Output}  \\
        \hline 
        \textbf{CNN} & \\
         Input layer (image patch) & $128{\times}128{\times}1$          \\
         Cv [$9{\times}9$] - ReLU - BN - AP [$2{\times}2$] - Dr ($10\%$)      & $64{\times}64{\times}32$\\
         Cv [$7{\times}7$] - ReLU - BN - AP [$2{\times}2$] - Dr ($20\%$)      & $32{\times}32{\times}64$\\
         Cv [$5{\times}5$] - ReLU - BN - Dr ($30\%$)      & $32{\times}32{\times}128$ \\
         Cv [$5{\times}5$] - ReLU - BN - AP [$2{\times}2$] - Dr ($30\%$)       & $16{\times}16{\times}128$\\
         Cv [$3{\times}3$] - ReLU - BN  - Dr ($40\%$)      & $16{\times}16{\times}256$ \\
         Cv [$3{\times}3$] - ReLU - BN -  GAP - Dr ($40\%$) & $256$ \\           
         \hline 
          \textbf{DNN} & \\
         Input layer (vector) & $256$ \\
         FC - ReLU - Dr ($50\%$)        &  $512$       \\
         FC - ReLU - Dr ($50\%$)        &  $1024$    \\
         FC - Softmax     &  10        \\
       \hline 
    \end{tabular}
    }
    \label{table:CDNN} 
\end{table}
\subsection{Experimental Setting}
Similar to experiments with MNIST dataset and MLP based network, the CNN based network is also built on Tensorflow framework, use Cross-entropy for loss function as mentioned in (\ref{eq:loss_func}), and Adam for optimization. 

\subsection{Experimental Results and Discussion}
\begin{table}[h]
    \caption{Performance over DCASE 2018 Task 1A} 
    \centering
    \scalebox{0.9}{

    \begin{tabular}{l c c} 
        \hline 
            \textbf{Network Option}   &  \textbf{Activation Func.} &  \textbf{Acc. \%}  \\
        \hline 
         w/o BN, w/o Dr & ReLU  &  53.2    \\
         w/o BN, w/o Dr& ELU  &  54.7        \\
         w/o BN, w/o Dr& SELU  &  57.1        \\
         w/o BN, w/o Dr& GELU  &  58.5        \\
         w/o BN, w/o Dr& ISRLU  &  57.3        \\
        \hline 
         \textbf{w/  BN, w/ Dr}& \textbf{ReLU}  &  \textbf{63.9}        \\ w/ BN, w/ Dr& ELU  &  61.7        \\
         w/ BN, w/ Dr& SELU  &  61.4        \\
         w/ BN, w/ Dr& GELU  &  63.1        \\
         w/ BN, w/ Dr& ISRLU  &  60.1        \\        
                 \hline 
         w/o BN, w/ Dr& SELU  &  59.1        \\
         w/ BN, w/o Dr& GELU  &  60.0        \\
       \hline 
    \end{tabular}
    }
    \label{table:result} 
\end{table}
To evaluate activation functions with the CNN based network architecture and DCASE 2018 Task 1A dataset, we propose to conduct three main experiments. 
In the first experiment, we uses the CNN based network proposed, but remove Batch normalization and Dropout layers. Next, ReLU is replaced by the other activation functions to evaluate.
The second experiment keeps the CNN based network proposed unchanged, thus replace ReLU by the other activation functions.
For the final experiment, we evaluate whether SELU and GELU can replace the roles of Bach normalization and Dropout layers as these activation functions' advantages mentioned in \ref{act_func}. To evaluate SELU, we remove the Batch normalization layer in the CNN based network. Meanwhile, we remove the Dropout layer in the CNN based network to evaluate GELU. 

As results shown in the top part of Table \ref{table:result}, without using Batch normalization and Dropout layers, activation functions such as ELU, SELU, GELU and ISRLU help to improve the classification accuracy compared with ReLU. This is similar to experiments' result over MNIST dataset and MLP based network mentioned in Section \ref{mpl}.
Again, using activation functions such as ELU, SELU, GELU and ISRLU to deal with \textit{vanished gradients} issue is confirmed.

As results shown in the middle part of Table \ref{table:result}, when Batch normalization and Dropout layers are used, CNN based network with ReLU outperforms the other activation functions. It is interesting to indicate that Batch normalization and Dropout layers effectively help to improve performance, thus reduce the effect of \textit{vanish gradient} from using ReLU.

Looking at the bottom part in Table \ref{table:result}, it can be seen that SELU and GELU cannot replace the roles of Batch normalization and Dropout layers within the experiment mentioned (CNN based network over DCASE 2018 Task 1A). Compare with DCASE 2018 Task 1A  baseline~\cite{Mesaros2018_DCASE}, our best result of 63.9\% improves by 4.2\%. As this paper focuses on analysing activation functions rather than competing with systems submitted to DCASE 2018 Task 1A challenge, the proposed model only uses single spectrogram and compared with the DCASE 2018 Task 1A baseline (Almost submitted systems for this challenge or papers~\cite{lampham_01, lampham_02, lampham_03} exploring this dataset make use of multiple spectrograms). 

\section{Conclusion}
This paper has presented an analysis of using different activation functions in deep learning networks with respect to supervised learning. By conducting extensive experiments on MLP based network with MNIST dataset and CNN based network with DCASE 2018 Task 1A dataset, the roles of activation functions are confirmed, thus further evaluated over the specific application of Acoustic Scene Classification. The experiments indicate that although ReLU, ELU, SELU, GELU and ISRLU can help to prevent \textit{vanished gradients} or/and \textit{dead state} issue, these activation functions should combine with other statistic layers such as Batch normalization or Dropout to further improve the ASC system performance.   

\addtolength{\textheight}{-12cm}   

\bibliographystyle{IEEEbib}
\bibliography{refs}

\begin{thebibliography}{10}

\bibitem{mohammadzaheri2009combination}
Morteza Mohammadzaheri, Lei Chen, Ali Ghaffari, and John Willison,
\newblock ``A combination of linear and nonlinear activation functions in
  neural networks for modeling a de-superheater,''
\newblock {\em Simulation Modelling Practice and Theory}, vol. 17, no. 2, pp.
  398--407, 2009.

\bibitem{kalman1992tanh}
Barry~L Kalman and Stan~C Kwasny,
\newblock ``Why tanh: choosing a sigmoidal function,''
\newblock in {\em [Proceedings 1992] IJCNN International Joint Conference on
  Neural Networks}. IEEE, 1992, vol.~4, pp. 578--581.

\bibitem{han1995influence}
Jun Han and Claudio Moraga,
\newblock ``The influence of the sigmoid function parameters on the speed of
  backpropagation learning,''
\newblock in {\em International Workshop on Artificial Neural Networks}.
  Springer, 1995, pp. 195--201.

\bibitem{krizhevsky2017imagenet}
Alex Krizhevsky, Ilya Sutskever, and Geoffrey~E Hinton,
\newblock ``Imagenet classification with deep convolutional neural networks,''
\newblock {\em Communications of the ACM}, vol. 60, no. 6, pp. 84--90, 2017.

\bibitem{clevert2015fast}
Djork-Arn{\'e} Clevert, Thomas Unterthiner, and Sepp Hochreiter,
\newblock ``Fast and accurate deep network learning by exponential linear units
  (elus),''
\newblock {\em arXiv preprint arXiv:1511.07289}, 2015.

\bibitem{selu}
G{\"u}nter Klambauer, Thomas Unterthiner, Andreas Mayr, and Sepp Hochreiter,
\newblock ``Self-normalizing neural networks,''
\newblock in {\em Advances in neural information processing systems}, 2017, pp.
  971--980.

\bibitem{hendrycks2016gaussian}
Dan Hendrycks and Kevin Gimpel,
\newblock ``Gaussian error linear units (gelus),''
\newblock {\em arXiv preprint arXiv:1606.08415}, 2016.

\bibitem{srivastava2014dropout}
Nitish Srivastava, Geoffrey Hinton, Alex Krizhevsky, Ilya Sutskever, and Ruslan
  Salakhutdinov,
\newblock ``Dropout: a simple way to prevent neural networks from
  overfitting,''
\newblock {\em The journal of machine learning research}, vol. 15, no. 1, pp.
  1929--1958, 2014.

\bibitem{gelu}
Dan Hendrycks and Kevin Gimpel,
\newblock ``Gaussian error linear units (gelus),''
\newblock {\em arXiv preprint arXiv:1606.08415}, 2016.

\bibitem{isrlu}
Brad Carlile, Guy Delamarter, Paul Kinney, Akiko Marti, and Brian Whitney,
\newblock ``Improving deep learning by inverse square root linear units
  (isrlus),''
\newblock {\em arXiv preprint arXiv:1710.09967}, 2017.

\bibitem{Goodfellow-et-al-2016}
Ian Goodfellow, Yoshua Bengio, and Aaron Courville,
\newblock {\em Deep Learning},
\newblock MIT Press, 2016,
\newblock \url{http://www.deeplearningbook.org}.

\bibitem{lecun1998gradient}
Yann LeCun, L{\'e}on Bottou, Yoshua Bengio, and Patrick Haffner,
\newblock ``Gradient-based learning applied to document recognition,''
\newblock {\em Proceedings of the IEEE}, vol. 86, no. 11, pp. 2278--2324, 1998.

\bibitem{ba2014deep}
Jimmy Ba and Rich Caruana,
\newblock ``Do deep nets really need to be deep?,''
\newblock in {\em Advances in neural information processing systems}, 2014, pp.
  2654--2662.

\bibitem{kingma2014adam}
Diederik~P Kingma and Jimmy Ba,
\newblock ``Adam: A method for stochastic optimization,''
\newblock {\em arXiv preprint arXiv:1412.6980}, 2014.

\bibitem{Mesaros2018_DCASE}
Annamaria Mesaros, Toni Heittola, and Tuomas Virtanen,
\newblock ``A multi-device dataset for urban acoustic scene classification,''
\newblock in {\em Proceedings of the Detection and Classification of Acoustic
  Scenes and Events 2018 Workshop (DCASE2018)}, November 2018, pp. 9--13.

\bibitem{lyon2017human}
Richard~F Lyon,
\newblock {\em Human and machine hearing},
\newblock Cambridge University Press, 2017.

\bibitem{mcfee2015librosa}
Brian McFee, Colin Raffel, Dawen Liang, Daniel~PW Ellis, Matt McVicar, Eric
  Battenberg, and Oriol Nieto,
\newblock ``librosa: Audio and music signal analysis in python,''
\newblock in {\em Proceedings of the 14th python in science conference}, 2015,
  vol.~8, pp. 18--25.

\bibitem{simonyan2014very}
Karen Simonyan and Andrew Zisserman,
\newblock ``Very deep convolutional networks for large-scale image
  recognition,''
\newblock {\em arXiv preprint arXiv:1409.1556}, 2014.

\bibitem{lampham_01}
Lam Pham, Huy Phan, Truc Nguyen, Ramaswamy Palaniappan, Alfred Mertins, and Ian
  McLoughlin,
\newblock ``Robust acoustic scene classification using a multi-spectrogram
  encoder-decoder framework,''
\newblock {\em arXiv preprint arXiv:2002.04502}, 2020.

\bibitem{lampham_02}
Lam Pham, Ian McLoughlin, Huy Phan, Ramaswamy Palaniappan, and Yue Lang,
\newblock ``Bag-of-features models based on c-dnn network for acoustic scene
  classification,''
\newblock in {\em Audio Engineering Society Conference: 2019 AES International
  Conference on Audio Forensics}. Audio Engineering Society, 2019.

\bibitem{lampham_03}
Dat Ngo, Hao Hoang, Anh Nguyen, Tien Ly, and Lam Pham,
\newblock ``Sound context classification basing on join learning model and
  multi-spectrogram features,''
\newblock {\em arXiv preprint arXiv:2005.12779}, 2020.

\end{thebibliography}


\end{document}